\documentclass[runningheads,a4paper]{llncs}

\usepackage{amssymb}
\usepackage{subfig}
\usepackage{cite}
\usepackage{graphicx}
\usepackage{algorithm}
\usepackage{algpseudocode}
\usepackage{url}
\usepackage{amsmath}
\usepackage{setspace}
\usepackage{multirow}
\usepackage{booktabs}

\usepackage{url}
\urldef{\mailsa}\path|{alfred.hofmann, ursula.barth, ingrid.haas, frank.holzwarth,|
\urldef{\mailsb}\path|anna.kramer, leonie.kunz, christine.reiss, nicole.sator,|
\urldef{\mailsc}\path|erika.siebert-cole, peter.strasser, lncs}@springer.com|
\newcommand{\keywords}[1]{\par\addvspace\baselineskip
\noindent\keywordname\enspace\ignorespaces#1}

\begin{document}

\mainmatter  

\title{Unsupervised Learning-based Depth Estimation aided Visual SLAM Approach}

\titlerunning{Lecture Notes in Computer Science: Authors' Instructions}

%
%

\author{Mingyang Geng $^{1}$ 
	\and Suning Shang $^1$\and Bo Ding $^{1*}$  \and Huaimin Wang $^1$\and Pengfei Zhang $^1$\and Lei Zhang $^2$}
\authorrunning{Mingyang Geng $^{1}$ \and  Suning Shang $^1$ \and  Bo Ding $^{1*}$ et al.}


\institute{National Key Laboratory of Parallel and Distributed Processing,\\
	College of Computer, National University of Defense Technology, China $^1$\\
	National Key Laboratory of Integrated Automation of Process Industry,\\ Northeastern University, China $^2$ 	
}

%
%

\toctitle{Unsupervised Learning-based Depth Estimation aided Visual SLAM Approach}
\tocauthor{Authors' Instructions}
\maketitle

\begin{abstract}

Existing visual-based SLAM systems mainly utilize the three-dimensional environmental depth information from RGB-D cameras to complete the robotic synchronization localization and map construction task. However, the RGB-D camera maintains a limited range for working and is hard to accurately measure the depth information in a far distance. Besides, the RGB-D camera will easily be influenced by strong lighting and other external factors, which will lead to a poor accuracy on the acquired environmental depth information. Recently, deep learning technologies have achieved great success in the visual SLAM area, which can directly learn high-level features from the visual inputs and improve the estimation accuracy of the depth information. Therefore, deep learning technologies maintain the potential to  extend the source of the depth information and improve the performance of the SLAM system. However, the existing deep learning-based methods are mainly supervised and require a large amount of ground-truth depth data, which is hard to acquire because of the realistic constraints. In this paper, we first present an unsupervised learning framework, which not only uses image reconstruction for supervising but also exploits the pose estimation method to enhance the supervised signal and add training constraints for the task of monocular depth and camera motion estimation. Furthermore, we successfully exploit our unsupervised learning framework to assist the traditional ORB-SLAM system when the initialization module of ORB-SLAM method could not match enough features. Qualitative and quantitative experiments have shown that our unsupervised learning framework performs the depth estimation task comparably to the supervised methods and outperforms the previous state-of-the-art approach by $13.5\%$ on KITTI dataset. Besides, our unsupervised learning framework could significantly accelerate the initialization process of ORB-SLAM system and effectively improve the accuracy on environmental mapping in strong lighting and weak texture scenes.     

\end{abstract}

\section{Introduction}

Simultaneous Localization and Mapping (SLAM) has attracted increasing attention in the robotic areas. SLAM technologies have wide applications in area such as autonomous driving, localization and navigation. The goal of a SLAM system is to construct the map of an unknown environment incrementally based on the perception information, i.e., scene information acquired by a radar or depth sensor when the robot is performing a complex task and confronted with an unknown environment. In order to achieve a satisfying performance in the visual SLAM tasks, the quality of the perception on the environmental depth, i.e., the distance
of the objects in the environment, will play an indispensable role. Therefore, how to extract valuable depth information from the visual inputs is an important problem in the visual SLAM systems.

Existing visual-based SLAM systems mainly utilize the three-dimensional environmental depth information from RGB-D cameras. However, the RGB-D camera maintains a limited range for working and is hard to accurately measure the depth information in a far distance. Besides, in some special scenes, i.e., strong lighting and weak texture environments, robotic visual SLAM always faces the problems of scale drift or scale error because of the inaccurate accuracy on the acquired depth information. The reason of obtaining the imprecise depth information is that most of the existing visual SLAM algorithms design sparse image features manually, while the manually designed features often contain certain assumptions about the environment, i.e., sufficient illumination, material determination, which will lead to a poor performance on environmental depth estimation when the environmental factors change. 

\begin{figure}[htbp]
	\centering
	\includegraphics[width=0.9\textwidth]{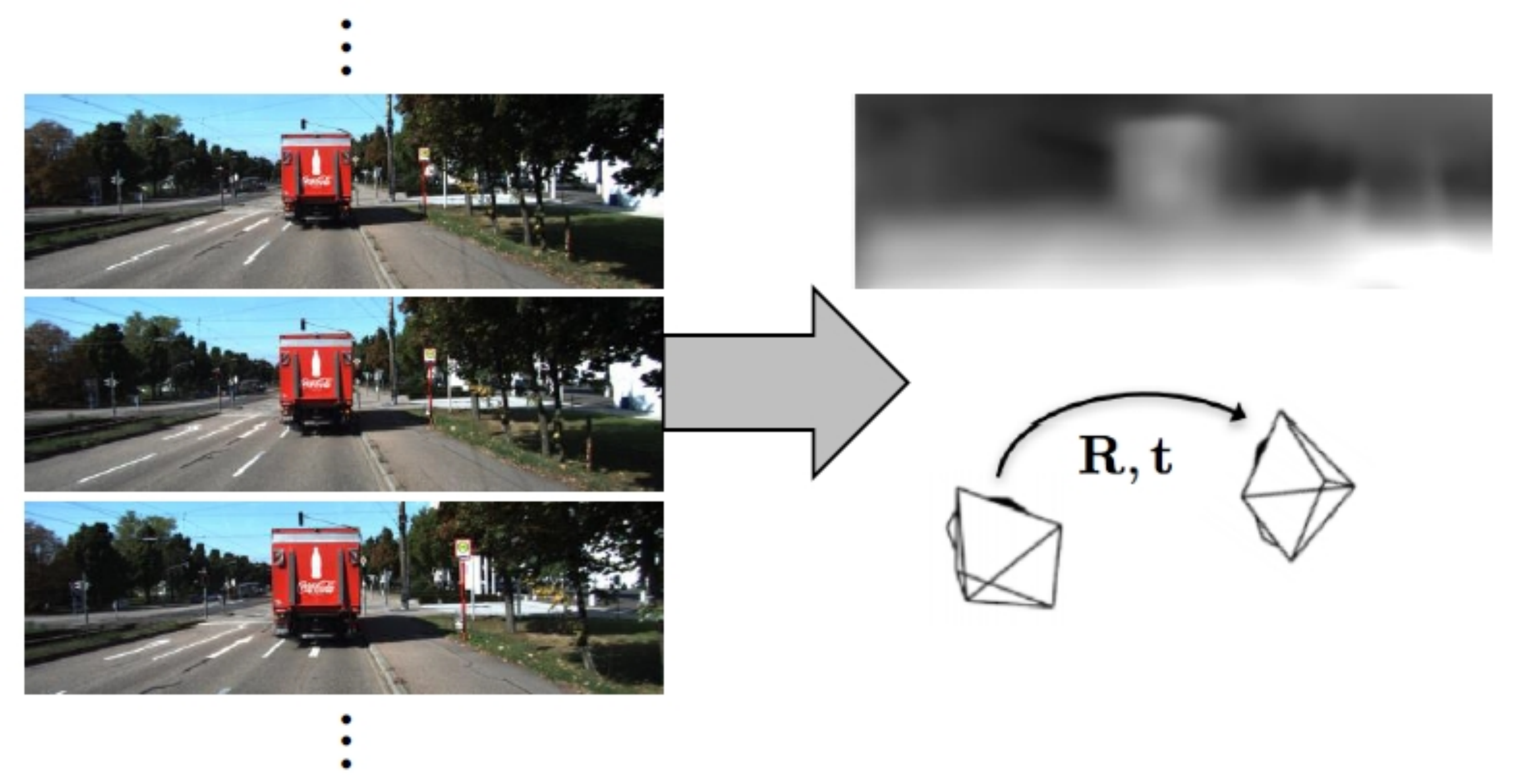}
	\caption{Illustration of our learning framework. The input to our system
		consists solely of unlabeled video clips. Our learning framework
		estimates the depth in the first image and the camera motion.}
\end{figure} 

Deep learning technologies have recently emerged as a powerful tool for improving the accuracy on monocular depth estimation \cite{eigen2014depth,liu2016learning,godard2017unsupervised,zhou2017unsupervised}. One of the advantages of deep learning technologies over common alternatives is that features are learned directly from data, and do not have to be chosen or designed by the algorithm developers for the specific problem
on which they are applied. However, existing methods \cite{eigen2014depth,liu2016learning,godard2017unsupervised} are mainly supervised and need a large amount of ground-truth data, which is hard to acquire because of the expensive radar sensors and the limited working range. A promising branch in the depth estimation field is unsupervised learning \cite{zhou2017unsupervised}, which exploits image reconstruction as supervision signal and significantly reduce the burden to collect high-quality depth training data in advance. However, the existing unsupervised method \cite{zhou2017unsupervised} does not fully exploit the heuristic knowledge during the image acquisition process, which can strengthen the supervised signal and further improve the accuracy on depth estimation. Therefore, how to enhance the supervised signal and utilize the unsupervised learning technologies to assist the traditional visual SLAM systems remains a great challenge.
 
In this paper, we first propose a novel unsupervised learning
framework which exploit the pose estimation method to enhance the supervised signals and further promote the accuracy of extracting the
depth information from monocular image sequences. In concrete, we utilize a large number of scene image sequences to train a model for camera motion
prediction and scene structure prediction (shown in Fig. 1). In the pose estimation stage, we set up a continuous frame window
and exploit the pose transformation
relationships to construct the pose graph, which can partially eliminate the cumulative error. Furthermore, we successfully exploit our unsupervised learning framework to assist the traditional ORB-SLAM system, a widely used visual SLAM system, when the initialization module of ORB-SLAM method could not match enough features. Extensive experiments have shown that our method can significantly accelerate the initialization process of ORB-SLAM system and effectively improve the accuracy on environmental mapping in strong lighting and weak texture scenes.

The rest of this paper is organized as follows. Section 2 introduces the background and the highly related work.
Section 3 describes the methodology of our work as well
as the architecture designed for training and prediction. Section 4 describes the details of our unsupervised learning-based depth estimation-aided visual SLAM system. The
validation and evaluation of our work based on different public
datasets are described in Section 5. Section 6 presents the experimental results implemented in the real-world settings. We conclude and provide
our future direction in Section 7. 

\section{Related Work}

Our method covers three research areas, including depth estimation optimization, monocular depth estimation and motion estimation from images. 
 
\subsection{Depth Estimation optimization}

The existing mechanism of depth estimation optimization can be mainly divided into three categories based on how to utilize the deep learning technologies. 

The first kind of methods directly substitute the depth information acquisition module with deep learning technologies \cite{gao2017unsupervised}. This method can effectively suit for the environments which are hard for traditional visual SLAM methods to deal with, i.e., strong lighting and weak texture environments. However, this method will require a more complex system with a stronger computing and processing ability. In addition, deep learning technologies are easily over-fit to the dataset and will lead to a poor performance in the unfamiliar environments. Last but not least, there does not exist a method which fully substitute the depth information acquisition module with deep learning technologies in realistic applications. Therefore, this kind of method needs further validation.

The second kind of methods exploit the environmental depth information from the deep learning technologies and the traditional SLAM system simultaneously \cite{li2018undeepvo}. This method can optimize the environmental depth information used by the visual SLAM system and indirectly improve the accuracy on mapping and localization. However, this method needs to implement the two depth information acquisition methods simultaneously and require a stronger ability of computing and processing. In addition, the complex evaluation process to select the optimal depth information needs to be accurately implemented, which is hard to satisfy the quality of service. 

The third kind of methods complement the drawbacks of the deep learning technologies and the depth information acquisition module of traditional SlAM algorithm \cite{tateno2017cnn}. In concrete, deep learning technologies is only applied when the traditional SLAM algorithms can not obtain high-accuracy depth information. This method can effectively improve the accuracy of the depth estimation while maintaining the ability to guarantee the QoS requirement. There are very limited works which exploit the deep learning technologies to acquire the environmental depth estimation. A deep learning-aided LSD-SLAM algorithm is proposed in \cite{engel2014lsd}, which achieves a better result than the traditional LSD-SLAM algorithm and a stronger adaptability to the strong lighting and weak texture environments. We choose ORB-SLAM as our baseline method, which has wide applications in visual SLAM area but performs an unsatisfied performance in strong lighting and weak texture environments. To the best of our knowledge, this is the first work which combines the deep learning technologies and the ORB-SLAM system.

\subsection{Monocular Depth Estimation}

Monocular depth estimation is a basic low-level challenge problem which has been studied for decades. Early works on depth estimation using RGB images usually relied on hand-crafted features and probabilistic graphical models. \cite{hoiem2005automatic} introduced photo pop-up, a fully automatic method for creating a basic 3D model from a single photograph. In \cite{karsch2016depth}, the authors design Depth Transfer, a non-parametric approach where the depth of an input image is reconstructed by transferring the depth of multiple similar images and then applying some warping and optimizing procedures. Delage et al. in \cite{delage2006dynamic} proposed a dynamic Bayesian framework for recovering 3D
information from indoor scenes. A discriminatively-trained multi-scale Markov Random Field (MRF) was introduced in \cite{saxena20083}, in order to optimally fuse local and global features. Depth estimation is considered as an inference problem in a discrete-continuous
CRF in \cite{liu2014discrete}.

More recent approaches for depth estimation are based on
convolutional neural network(CNN). As a pioneer work, Eigen
et al. proposed a multi-scale approach for depth prediction
in \cite{eigen2014depth}. It considers two deep networks, one performing a
coarse global prediction based on the entire image, and the
other refining predictions locally. This approach was extended
in \cite{eigen2015predicting} to handle multiple tasks (e.g. semantic segmentation,
surface normal estimation). In \cite{liu2016learning}, authors combine a deep
CNN and a continuous conditional random field, and attain
visually sharper transitions and local details. In \cite{laina2016deeper}, a deep residual network is developed, based on the ResNet and
achieved higher accuracy than \cite{liu2016learning}. Unlike our approach,
these methods require explicit depth for training. Unsupervised
learning setups have also been explored for disparity image
prediction. For instance, Godard et al. formulate disparity
estimation as an image reconstruction problem in \cite{godard2017unsupervised}, where
neural networks are trained to warp left images for matching
the right one. Though these methods show similarity with
ours, which are unsupervised without requiring ground-truth
depth data for training, they assume camera poses known in advance, which is treated a large simplification. Our work is
inspired by that of \cite{zhou2017unsupervised}, which proposes to use view synthesis as
the supervisory signal. However, the further advantage of our
approach which demonstrated in the following evaluations is
that, the idea of continuous frame window used in traditional
SLAM approach is applied to enhance the supervisory signal
and capture more constraints which can guide the training
process for more accuracy results.

\subsection{Motion Estimation from Images}

The motion estimation has a long history in computer vision.
The underlying 3D geometry is a consolidated field. They
consist of a long pipeline of methods, start from descriptor
matching for finding a sparse set of correspondences between
images \cite{lowe2004distinctive}, to estimating the essential matrix to determine the
camera motion. Bundle adjustment \cite{triggs1999bundle} is used in the pipeline
of method to refine the final structure and camera position.
The bundle adjustment minimizes the reprojection error of
the three-dimensional point in the two-dimensional image
sequence by Levenberg-Marquardt (LM) nonlinear algorithm
to get the optimal motion model \cite{ lourakis2005brief}. The accuracy of the
bundle adjustment method is related to the number of frames
of the image. The more the number of image frames, the more
accurate the camera motion parameters can obtain.

Recent works \cite{dosovitskiy2015flownet} propose learning frame-to-frame motion
fields with deep neural networks supervised with ground-truth
motion obtained from simulation or synthetic movies. This
enables efficient motion estimation that learns to deal with
lack of texture using training examples rather than relying
only on smoothness constraints of the motion field, as previous
optimization methods \cite{ sun2010secrets}. Our approach draws on the respective
advantages of the geometry-based motion estimation
in SLAM and the learning-based motion estimation. Multiview
pose network is used to estimate pose transformation
matrix between adjacent frames. We set up a continuous frame
window to construct the pose graph and use the pose transform
relationship to calculate more pose transform matrices which
perfect the pose graph.

\section{Pose Estimation-based Monocular Depth Estimation Method}

The accuracy of existing monocular depth estimation methods is hard to satisfy the requirement realistic applications. Therefore, it is meaningful to improve the accuracy of monocular depth estimation. In this section, we introduce our pose estimation-based monocular depth estimation method from the following three aspects: the basic framework, learning and geometry-based pose estimation and image recovery-based training method.

\subsection{Framework}

\begin{figure}[htbp]
	\centering
	\includegraphics[width=0.9\textwidth]{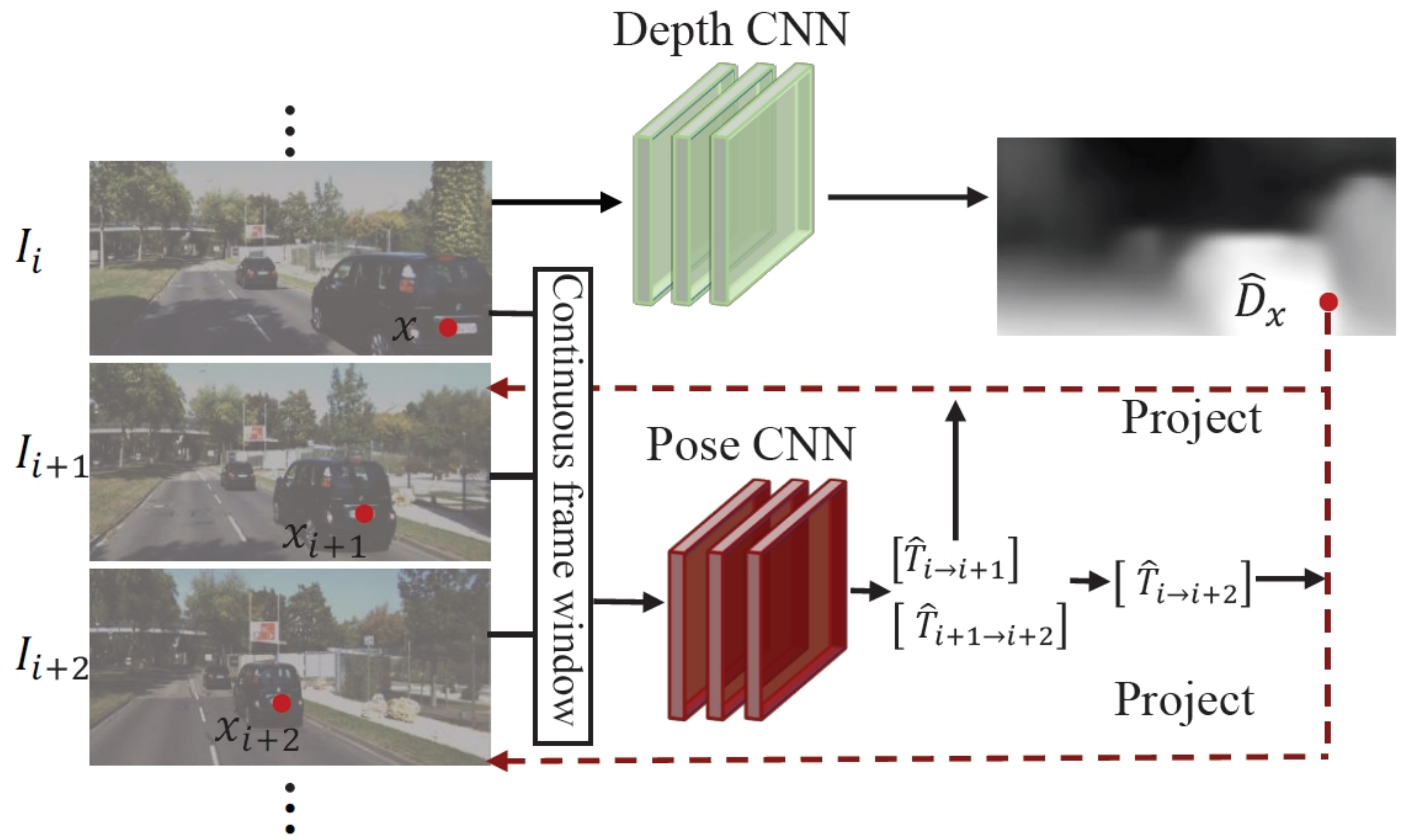}
	\caption{The overview of the training pipeline based on image
		reconstruction. The depth network takes only the first image $I_{i}$
		as input and outputs a per-pixel depth map $\hat{D}_{x}$. The continuous
		frame window takes images (e.g., Ii, Ii+1, Ii+2) as input, through the
		pose network, outputs the relative camera pose matrices of adjacent
		frames ( $\hat{T}_{i\rightarrow i+1}$, $\hat{T}_{i+1\rightarrow i+2}$) and we can use the camera pose matrices
		to calculate more camera poses ( $\hat{T}_{i+1\rightarrow i+2}$). The
		outputs of both models have then used to inverse warp images to reconstruct the target image, and the photometric
		reconstruction loss is used for training the CNNs. By utilizing image
		reconstruction as supervision, we are able to train
		the entire framework in an unsupervised manner from videos.}
\end{figure}

Given a single image frame $I$, the goal of our method is to provide two functions $f_{1}$ and $f_{2}$ which can predict the per-pixel scene depth $d^{'}=f_{1}I$ and the camera pose $p^{'}=f_{2}I$. We design two deep neural networks (depth estimation neural network and pose estimation neural network) to learn these two functions. Most existing methods treat the learning task as a supervised learning problem, where the color input images, the corresponding target depth and pose values are provided. However, it is not practical to acquire such large amount of ground-truth depth and pose data in various scenes because of the expensive lidar sensor and the limited working range. Besides, existing methods always neglect the traditional pose estimation methods and do not take the prior knowledge from traditional algorithms into account. 

\begin{figure}[htbp]
	\centering
	\includegraphics[width=0.9\textwidth]{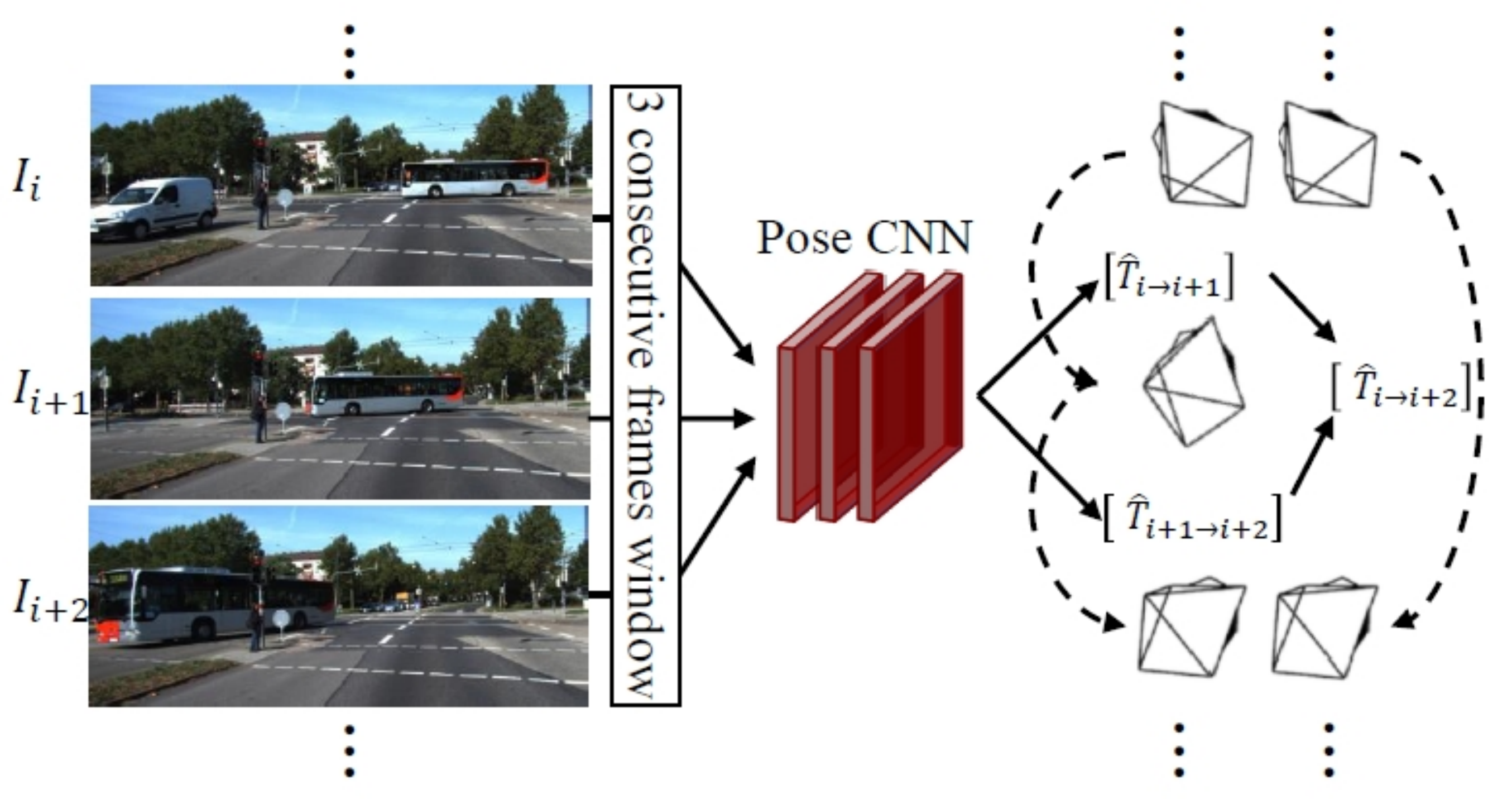}
	\caption{Illustration of the camera pose estimation pipeline. For
		example, we maintain a continuous frame window whose length is
		3 frames for estimating the camera pose transform matrix. For each sequence images (e.g., $I_{i},I_{i+1},I_{i+2}$), we will get the adjacent camera pose transform matrix $(\hat{T}_{i\rightarrow i+1},\hat{T}_{i+1\rightarrow i+2})$. We can use the camera
		pose transformation to get more camera poses $\hat{T}_{i\rightarrow i+2}$.}
\end{figure} 

We propose an unsupervised learning method, which exploits the pose estimation approach in traditional SLAM algorithms to augment the supervised signals by image reconstruction. In concrete, based on a short image sequences $I_{i},I_{i+1},I_{i+2}$captured by a moving camera, we can reconstruct the image $I_{i}^{'}$ by the predicted the depth image $D_{i}$ and the predicted pose estimation matrix. The difference between the image $I_{i}$ and the reconstructed image $I_{i}^{'}$ can be used as the supervised signals to train the depth estimation and the pose estimation neural networks. The framework of our unsupervised method is illustrated in Fig. 2. The monocular video sequences are used for training the single-view depth estimation and the multi-view pose estimation networks. The output of the single-view depth estimation network is the depth map of the input image. For the pose estimation part, a continuous frame window are used as the input and the output of the multi-view pose estimation network is the pose transform matrices between all adjacent frames in the continuous frame
window. We then optimize the pose graph in the continuous frame window by calculating more nonadjacent pose transform matrices using a pose transform relationship. Then, we can reconstruct the image by the depth map and the pose transform matrices and train the two neural networks by calculating the difference between the input and the reconstructed images. 

\subsection{Estimation based on Learning and Geometry Pose Graph}  	  
Given a frame $I_{i}$, the single-view depth estimation network can directly predict the corresponding depth map $\hat{D}_{i}$. For the pose estimation part, our method is based both on the unsupervised learning technologies and the traditional geometry pose graph. A continuous frame window is set up before the multi-view pose estimation network in order to make the network learn the pose relationship between the continuous images $<I_{1},I_{2},\cdots,I_{n}>$ simultaneously. The
length of the continuous frame window stands for the number of input images in a training episode. In other words, the continuous frame window sequentially reads $n$ images from the training set and then sends the training data to the multi-view pose estimation network for further processing. During the training process, the multi-view pose estimation network will sequentially predict the transformation matrix between two adjacent frames in the continuous frame window
(shown in Fig. 3). For a better illustration, denote the the input image sequences in the continuous frame window as $<I_{1},I_{2},\cdots,I_{n}>$, the output of the multi-view pose estimation network is $\hat{T}_{i\rightarrow i+1}$. Then, we can build a preliminary pose graph using the pose transformation matrices between the adjacent frames. However, the preliminary pose graph lacks the pose transform matrices between non-adjacent frames, so we calculate the non-adjacent pose transformation relationships by using the following function:

\begin{equation}
\hat{T}_{i\rightarrow i+1} \times \hat{T}_{i+1\rightarrow i+2}=\hat{T}_{i\rightarrow i+2}
\end{equation}
	
Similarly, we can increase the length of the frame window to get more non-adjacent pose transformation relationships (e.g., $\hat{T}_{i\rightarrow i+5}$) and improve the pose graph. The acquired pose transformation relationships maintain the following two advantages. First, the cumulative error is partially eliminated. In the continuous
frame window, the error of pose estimation between
adjacent frames will accumulate gradually. But if we use
the calculated pose matrix to reconstruct the image, we can sequentially adjust the parameters of the pose network and partially eliminate the cumulative error. The second advantage is that this mechanism can avoid the estimation errors between the frames which are far apart in the frame sequence. Experiments \cite{zhou2017unsupervised} have shown that learning-based methods could not predict a satisfying relationship between the two frames which maintain a far distance in the frame sequence. Our method solve this problem by calculating the far apart pose relationships based on the pose estimation of the adjacent frames. 


\subsection{Geometry-based Image Reconstruction} 

\begin{figure}[htbp]
	\centering
	\includegraphics[width=0.9\textwidth]{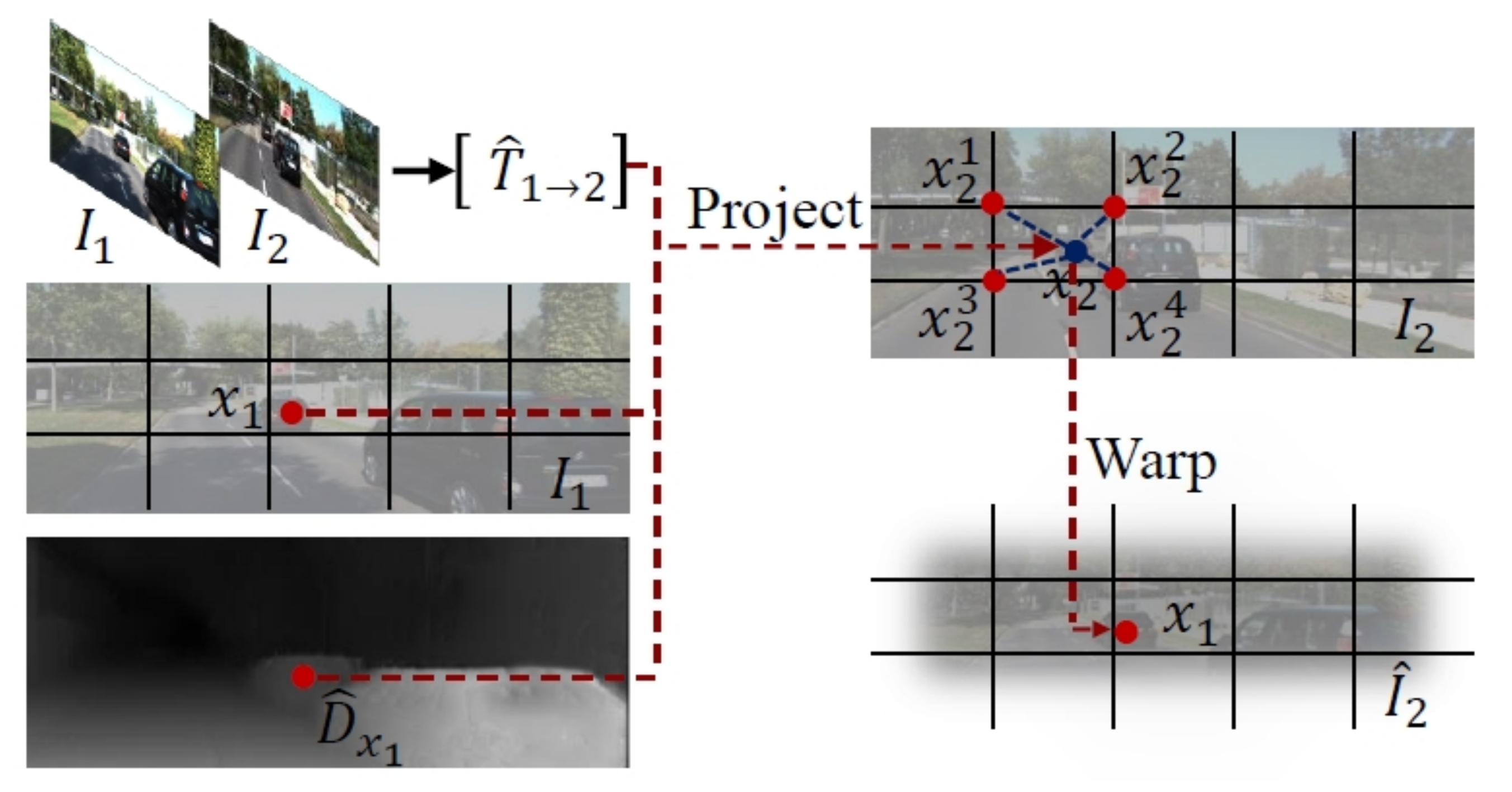}
	\caption{Illustration of image reconstruction based on camera pose
		matrix. For each point (e.g., $x_{1}$) in the first image, we project it onto the other image base on the predicted depth and camera pose
		and then use bilinear interpolation to obtain the value of the warped image $(\hat{I}_{2})$ at location ($x_{1}$).}
\end{figure} 

Image reconstruction through the means of warps and camera projection
is an important application of geometric scene understanding.
The goal of image reconstruction is to reconstruct a new
viewpoint’s image from other viewpoints’ through warps and
camera projection. In our learning framework, we
reconstruct the target image $I_{t}$ by sampling pixels from the other images $I_{r}$ based on the predicted depth map $\hat{D}_{t}$ and the predicted $4\times 4$ camera pose transformation matrix $\hat{T}_{t\rightarrow r}$. 

Our camera model is the pinhole model. Denote $K$ as the camera intrinsic matrix, $I_{1}$ and $I_{2}$, as the first and the second image in a training episodes. The transformation matrices of the two images to the world coordinates are represented as $T_{1 \rightarrow w}$ and $T_{2 \rightarrow w}$, and the homogeneous
coordinates of a pixel in the first image is represented as $x_{1}$. We can acquire the projected coordinates of $x_{1}$ onto the second image $x_{2}$ by $x_{2}\sim KT_{2 \rightarrow w}T_{2 \rightarrow w}^{-1}K^{-1}\hat{D}_{1}(x_{1})x_{1}$. Notice that the camera transformation matrix between $I_{1}$ and $I_{2}$ is equal to $T_{2 \rightarrow w}T_{1 \rightarrow w}^{-1}$, i.e., $\hat{T}_{1\rightarrow 2}=T_{2\rightarrow w}T_{1\rightarrow}^{-1}$. We substitute $T_{1\rightarrow w}T_{2 \rightarrow w}^{-1}$ with $\hat{T}_{1\rightarrow 2}$ so the formula becomes $x_{2}\sim K\hat{T}_{1 \rightarrow 2}K^{-1}\hat{D}_{1}(x_{1})x_{1}$. Furthermore,	
when we get a short image sequences $<I_{1},I_{2},\cdots,I_{n}>$ at the training time, denote $I_{t}$ as the target view image, and $I_{r}$ as the other images. The pixel project procedure can be formulated as:

\begin{equation}
x_{r}\sim K\hat{T}_{t \rightarrow r}K^{-1}\hat{D}_{t}(x_{t})x_{t}.
\end{equation}	 

Based on Eq.2, we
can project the pixels on the target image $I_{t}$ onto other images $I_{r}$.
After that, our image reconstruction model uses the image sampler from the spatial transformer network (STN) to sample the projected image $I_{r}$. The STN uses bilinear sampling where the output pixel is the weighted sum of the four pixel neighbors $x_{r}^{(1)},x_{r}^{(2)},x_{r}^{(3)},x_{r}^{(4)}$ of $x_{r}$, i.e., $\hat{I}_{r}(x_{t})=I_{r}(x_{r})=\sum_{i=1}^{4}w^{(i)}I_{r}(x_{r}^{i})$, where $w^{(i)}$ is linearly proportional to the spatial proximity between $x_{r}$ and $x_{r}^{i}$, and $\sum_{i=1}^{4}w^{(i)}=1$ (shown in Fig. 4). Contrast with the alternative approaches \cite{garg2016unsupervised},
the bilinear sampler is locally fully differentiable and
integrates seamlessly into our fully convolutional network, which means that we do not require any simplification or
approximation of our cost function. 

Finally, we use the predicted depth map $\hat{D}_{t}$ and the predicted $4\times 4$ camera pose transformation matrix $\hat{T}_{t\rightarrow r}$ of the previous step to reconstruct the target image through projection and the differentiable bilinear sampling mechanism.	

\subsection{Image Reconstruction as Supervision}

Image reconstruction has been used to learn end-to-end
unsupervised optical flow \cite{jason2016back}, disparity flow in a stereo rig \cite{godard2017unsupervised}
and video prediction \cite{patraucean2015spatio}. These methods reconstruct the images
by transforming the input based on depth maps or flow
fields. Our work considers dense structure estimation and uses
monocular videos to obtain the necessary self-supervision,
instead of static images. The depth information could also be predicted from a single
image supervised by photometric error \cite{garg2016unsupervised}. However, the methods above do not
infer camera pose transform or object motion and require
stereo pairs with known baseline in the training process. Our work estimates the camera motion between frames, which effectively removes the constraint that the ground-truth pose of the camera is known in the training process. 

In concrete, denote the first and the second images acquired from a moving monocular camera in chronological order as $I_{1}$ and $I_{2}$. Instead of directly predicting the depth and the camera pose of the first view image $I_{1}$, we use the second image $I_{2}$ to reconstruct $I_{1}$, which is illustrated in the previous section. Every pixel point $x$ in $I_{2}$ coordinates is warped to the target coordinate image. Let $x$ indexes over the pixel coordinate, and denote $\hat{I}_{2}$ as the second image $I_{2}$ warped to the first coordinate frame based on the image reconstruction process.		 	

When we use image reconstruction as a supervised signal,
the difference between the reconstructed image and the target
image can be calculated by $C=\sum_{x}|I_{1}(x)-I_{2}(x)|$. Similarly, denote the short image sequences as $<I_{1},I_{2},\cdots,I_{n}>$ in the training process, $I_{t}$ as the target view image, and $I_{r}$ as the other images. The image reconstructs procedure can be formulated as: 

\begin{equation}
C_{vr}=\sum_{r}\sum_{x}|I_{t}(x)-\hat{I}_{r}(x)|,
\end{equation}

where the supervisory signal $C_{vr}$ adjust the parameters of our depth estimation and pose estimation networks.

\subsection{Network Architecture and Training Loss}

\begin{figure}[htbp]
	\centering
	\includegraphics[width=0.9\textwidth]{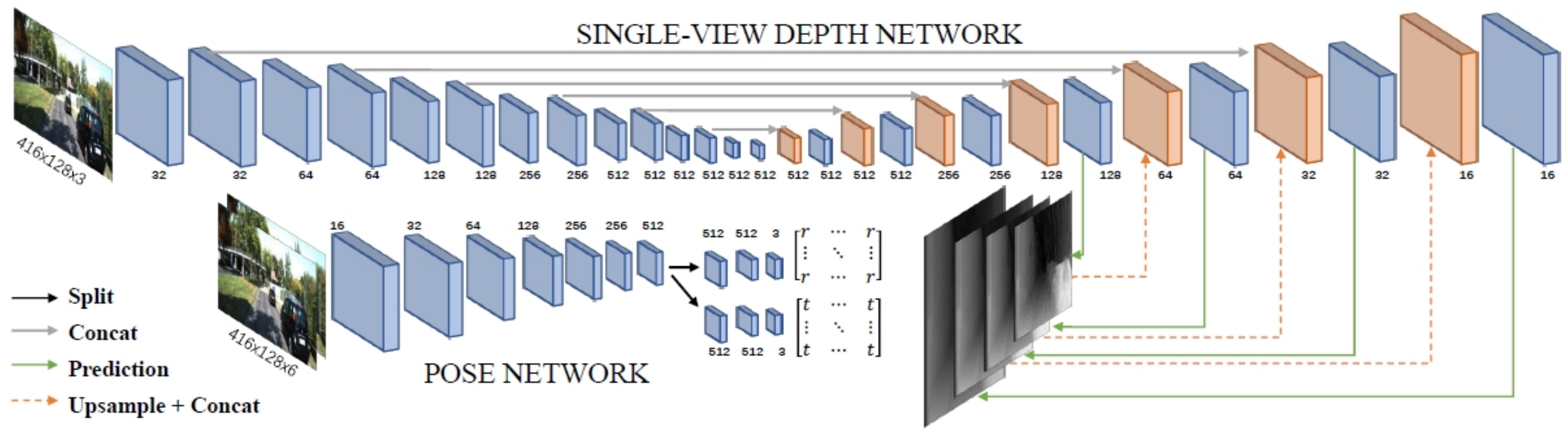}
	\caption{Network architecture for our depth and pose prediction modules. (a) For single-view depth network, we adopt the DispNet architecture
		with multi-scale side predictions. All conv layers are followed by ReLU activation except for the prediction layers. (b) For pose network,
		the input is sequences of consecutive frames. All conv layers are followed by ReLU except for the last layer where no nonlinear activation
		is applied.}
\end{figure}

As shown in Fig. 5, the
architecture consists of two different networks: the single-view
depth network and the pose network. The single-view depth
network is inspired by DispNet but several important
modifications are made to enable the training process without ground-truth
depth data. The single-view network is mainly composed of two parts: the encoder (from cnv1 to cnv7b) and the decoder
(from deconv7). The decoder uses skip connection layers to connect the activation blocks of the encoder, which enables the ability to obtain more representative features. The disparity are predicted at four
different scales (from disp4 to disp1). The function of
the pose network is to predict the relative poses between the
target image and other input images, which are accurately described by the 6-dimensional
camera pose transform matrix (3-dimensional euler angles and 3-dimensional translation).

In our learning framework, the gradients are mainly derived from the pixel intensity difference between the four-pixel neighbors of $x_{r}$ and $x_{t}$. Therefore, the training process will be inhibited when the correct $x_{r}$ (projected using
the ground-truth depth and pose) is located in a low-texture
region or far from the current estimation. We solve this problem by using multi-scale and smoothness loss which allows gradients to be derived from larger spatial regions directly.

Denote the loss at each output scale as $C_{s}$, so the total loss can be represented as  $C=\sum_{s=1}^{4}C_{s}$. Our loss module calculates $C_{s}$ as a combination of the two main terms: 

\begin{equation}
C_{s}=C_{vr}+\lambda C_{smooth},
\end{equation}

which encourages the reconstructed image to appear similar to the corresponding training input, indexes the minimized norm
of the second-order gradients for the predicted depth maps. $\lambda$ denotes the weighting for the depth smoothness loss. 

\section{Visual SLAM System with the Assistance of Unsupervised Learning-based Depth Estimation}

Existing visual SLAM system always acquires the depth information from the depth sensor. The depth sensor can directly obtain the environmental depth information within a certain distance. However, the depth sensors suffer from the limited working range and are sensitive to the interference, which will decrease the acquired accuracy. In Section 3, we propose an image depth estimation method based on camera pose transformation relationship, which can directly obtain the depth information from the image through unsupervised learning. In this section, we introduce our method which extends the source of the three-dimensional environmental depth information and improves the accuracy of the depth information on the basis of ORB-SLAM algorithm.

\subsection{Traditional ORB-SLAM Algorithm}

The ORB-SLAM algorithm is mainly composed of the following three parts: the tracking module, the local map building module and the closed-loop detection \cite{canutescu2003cyclic} module. The tracking module aims to locate the camera through each frame and determines whether the frame should join the key-frame set. The tracking task first extracts the features of the frame, then initializes the camera pose and map and tracks the local map, finally determines the key frame. The local map building module aims to process the new key-frames and rebuild the map through local BA \cite{agarwal2010bundle}. The closed-loop detection module mainly judges the newly added key-frame and determines whether the scene has been encountered before.

In the initialization phase of ORB-SLAM, the monocular camera first reads the RGB image, and the depth sensor acquires the depth data. Then, feature point matching process is performed on the two consecutive frames in the time series to determine the motion of the camera. The image and depth acquisition process is implemented until the number of matched feature points on the two consecutive frames reaches a specified threshold. Then, the ORB-SLAM algorithm utilizes the existing geometric relationship between the matched feature points to calculate the current pose of the camera, and further creates an initial map or updates the local map.

\subsection{Depth Estimation Optimizing Mechanism}

By analyzing the ORB-SLAM algorithm, we decide to optimize the depth information acquisition process in the initialization phase. In concrete, our unsupervised learning-based depth estimation mechanism is applied only when the ORB-SLAM system does not match enough feature points. In this way, the accuracy of the ORB-SLAM will be effectively optimized while maintaining a reasonable computing ability and satisfying the QoS requirement.  

The initialization phase of ORB-SLAM algorithm is optimized as follows. When there are not enough matched feature points between two consecutive frames, the RGB image will be transmitted to our monocular depth estimation network instead of making the system read the next frame. Then, our depth estimation network will re-estimate the environmental depth information of the input image and further implement the feature point matching process. The complete depth information optimization mechanism on the basis of ORB-SLAM algorithm is illustrated below: 

\begin{algorithm}[h] 
		\caption{The initialization process of the optimized ORB-SLAM algorithm using our monocular depth estimation network.} 
		\label{alg:Framwork}
		\begin{algorithmic}[1] 
			\Require 
			The threshold $M$, which stands for the number of the matched feature points required by the ORB-SLAM algorithm.  
			\State The monocular camera reads the RGB image and the depth sensor acquires the depth information.
			\State Feature point matching process is implemented on the two consecutive frames in time series based on the RGB image and the depth information.
			\If{the number of matched feature points is less than the threshold $M$}
			\State Transmit the current image to the monocular depth estimation network and get the new depth image.
			\State Implement the feature point matching process on the two consecutive frames in time series based on the RGB image and the new depth image.
			\If{the matched feature points still less than $M$}
			\State Read the next RGB image and get the corresponding depth information.
			\Else
			\State Get the pose of the camera by utilizing the geometric relationship between the matched feature points.
			\EndIf
			\Else
			\State Get the pose of the camera by utilizing the geometric relationship between the matched feature points.
			\EndIf
			\State Create the initial map or update the local map.
		\end{algorithmic} 
\end{algorithm}

\subsection{Implement of Depth Estimation Assisted Visual SLAM System based on Unsupervised Learning}

We now introduce the details to realize the optimized ORB-SLAM system using our unsupervised learning-based depth estimation method. For the training process of our monocular depth estimation networks, the training dataset includes samples from various scenes (i.e., indoor scenes, outdoor scenes) as well as samples of weak texture and strong light. After the training process, we decoupled the trained network because the optimization mechanism only uses the depth estimation network. In addition, we establish the transmission channel of the ORB-SLAM initialization module and the trained depth estimation network. When the ORB-SLAM algorithm does not match enough feature points, the RGB image is transmitted to the depth estimation network, which keeps the running state to guarantee a immediate output.

\section{Evaluation on the Testing Set}

In this section, we evaluate the performance of our approach and make comparison with the existing methods on single-view depth and ego-motion estimation. We choose KITTI dataset as the test benchmark. To evaluate the cross-dataset generalization ability of our approach and demonstrate the superiority on the strong lighting and weak texture environments, we also use the Make3D dataset for a better illustration.

\subsection{Training Details}

We implement our system in TensorFlow \cite{abadi2016tensorflow}. In all experiments, the value of $\lambda$ is set to 0.5. During the training process, we use the Adam optimizer \cite{kingma2014adam} with $\beta _{1}$ of 0.9, learning rate of 0.0002 and mini-batch size of 4. All the experiments are performed with image sequences captured with a monocular camera and the images are resized to $128\times 416$ during training. In the test phase, the depth and pose
networks can be applied independently for images of arbitrary size.

\subsection{Single-view Depth Estimation}

We present results for the KITTI dataset \cite{geiger2012we} using two different test splits, to enable comparison to existing works. In its raw form, the dataset contains $42382$ images from $61$
scenes. The length of the continuous frame window is set to 3.

\begin{figure}[htbp]
	\centering
	\includegraphics[width=0.9\textwidth]{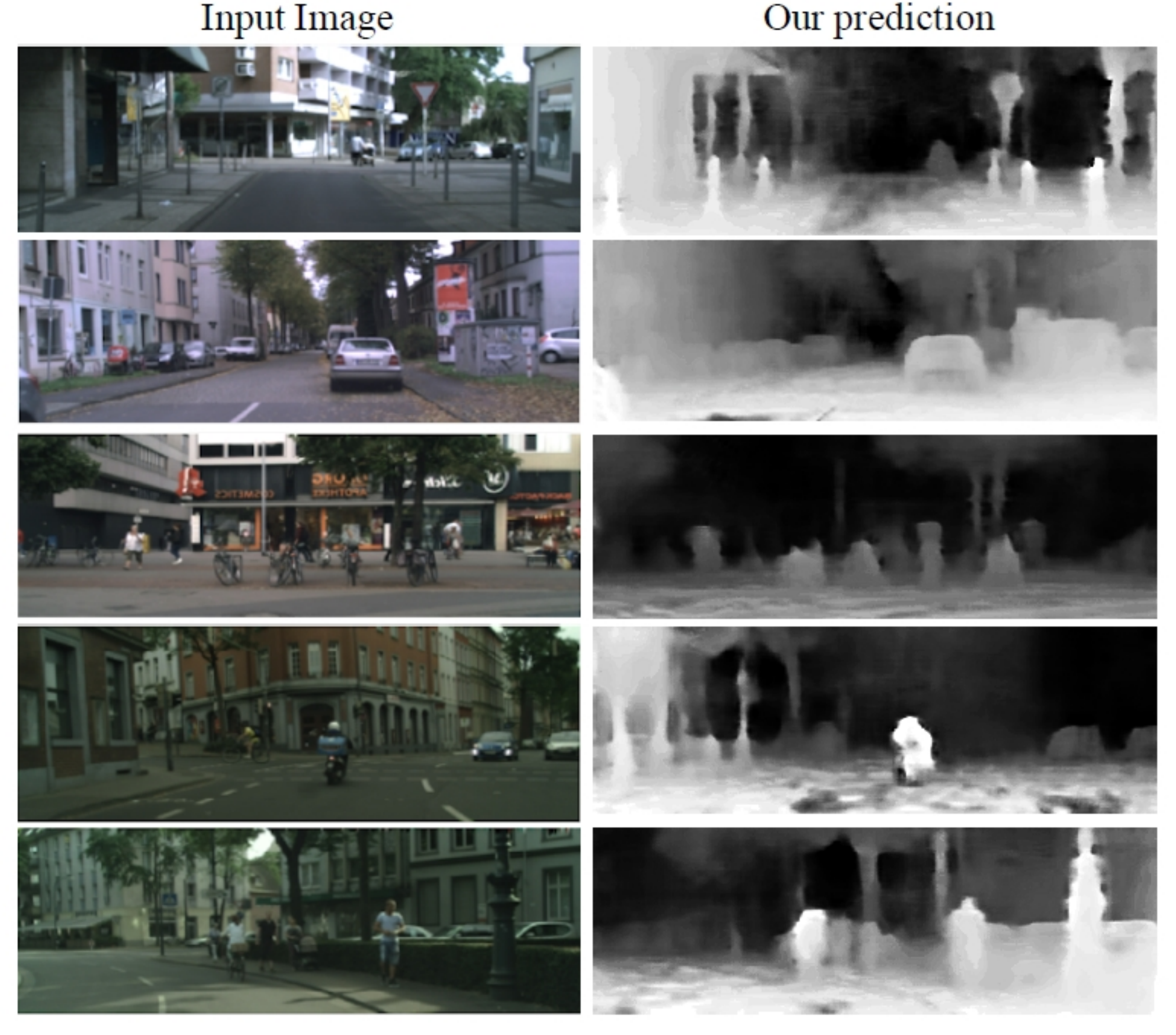}
	\caption{The sample prediction on Cityscapes dataset using our
		approach trained on Cityscapes only.}
\end{figure} 

To the best of our knowledge, among the methods which learn single-view depth estimation from monocular videos using unsupervised mechanism, state-of-the-art performance is achieved in Zhou \cite{zhou2017unsupervised}. We also make comparison with methods using supervised mechanism (depth ground-truth with depth supervision or calibrated stereo images with pose supervision) for training. Our method uses a scale factor to define the predicted depth so in the test phase, we multiply the predicted depth maps with a scalar which matches the median with the ground-truth data. Fig. 6 illustrates the predictions of our approach training
on Cityscapes dataset \cite{cordts2016cityscapes}. We also make comparison with Godard \cite{godard2017unsupervised} by taking the same training strategy which first pre-train the system on the Cityscapes dataset and then
fine-tune on the KITTI dataset.

\begin{table}[htbp] 
	\centering
	\caption{\label{tab:test}Single-view depth results on the KITTI dataset and Cityscapes dataset.} 
\setlength{\tabcolsep}{0.4mm}{
\begin{tabular}{lcccccccccc}
	\hline
	\multirow{2}{*}{Method}&
	\multirow{2}{*}{Dataset}&
	\multicolumn{2}{c}{Supervision}&
	\multicolumn{4}{c}{Error metric}&
	\multicolumn{3}{c}{Accuracy metric}\\
	\cmidrule(lr){3-4} \cmidrule(lr){5-8} \cmidrule(lr){9-11}
	~&~&Depth&Pose&Abs Rel&Sq Rel&RMSE&RMSE log&$\delta<1.25$&$\delta<1.25^2$&$\delta<1.25^3$\\
	\hline
	Train set mean&K&$\checkmark$&~&0.403&5.530&8.709&0.403&0.593&0.776&0.878\\
	Eigen $et\ al.$ Coarse&K&$\checkmark$&~&0.214&1.605&6.563&0.292&0.673&0.884&0.957\\
	Eigen $et\ al.$ Fine&K&$\checkmark$&~&0.203&1.548&6.307&0.282&0.702&0.890&0.958\\
	Liu $et\ al.$ &K&$\checkmark$&~&0.202&1.614&6.523&0.275&0.678&0.895&0.965\\
	Godard $et\ al.$ &K&$~$&\checkmark&0.148&1.344&5.927&0.247&0.803&0.922&0.964\\
	Godard $et\ al.$ &CS+K&$~$&\checkmark&0.124&1.076&5.311&0.219&0.847&0.942&0.973\\
	Zhou $et\ al.$ &K&$~$&~&0.208&1.768&6.856&0.283&0.678&0.885&0.957\\
	Zhou $et\ al.$ &CS+K&$~$&~&0.198&1.836&6.565&0.275&0.718&0.901&0.960\\
	$\textbf{Ours}$ &K&$~$&~&$\textbf{0.180}$&$\textbf{1.510}$&$\textbf{6.349}$&$\textbf{0.256}$&$\textbf{0.741}$&$\textbf{0.906}$&$\textbf{0.966}$\\
	$\textbf{Ours}$ &CS&$~$&~&$\textbf{0.236}$&$\textbf{2.476}$&$\textbf{7.249}$&$\textbf{0.307}$&$\textbf{0.645}$&$\textbf{0.861}$&$\textbf{0.946}$\\
	$\textbf{Ours}$ &CS+K&$~$&~&$\textbf{0.170}$&$\textbf{1.429}$&$\textbf{6.082}$&$\textbf{0.245}$&$\textbf{0.786}$&$\textbf{0.927}$&$\textbf{0.969}$\\
	\hline
\end{tabular}}
\end{table}

\subsubsection{KITTI}

We follow the experimental settings proposed by \cite{eigen2014depth} with the the test set of $697$ images covering $29$ scenes. Table 1 shows the performance comparison between our method and the baseline methods. Here, $K$ stands for the KITTI dataset and $CS$ stands for the $Cityscapes$ dataset. Compared with the methods using depth supervision \cite{eigen2014depth,liu2016learning}, our method performs better. However, our unsupervised method performs a little worse than the methods using pose supervision mechanism \cite{godard2017unsupervised,garg2016unsupervised}. \cite{godard2017unsupervised}
uses calibrated stereo images with left-right cycle consistency loss for training. In future work,
we will apply the similar cycle consistency loss to
our framework.

\begin{figure}[htbp]
	\centering
	\includegraphics[width=0.95\textwidth]{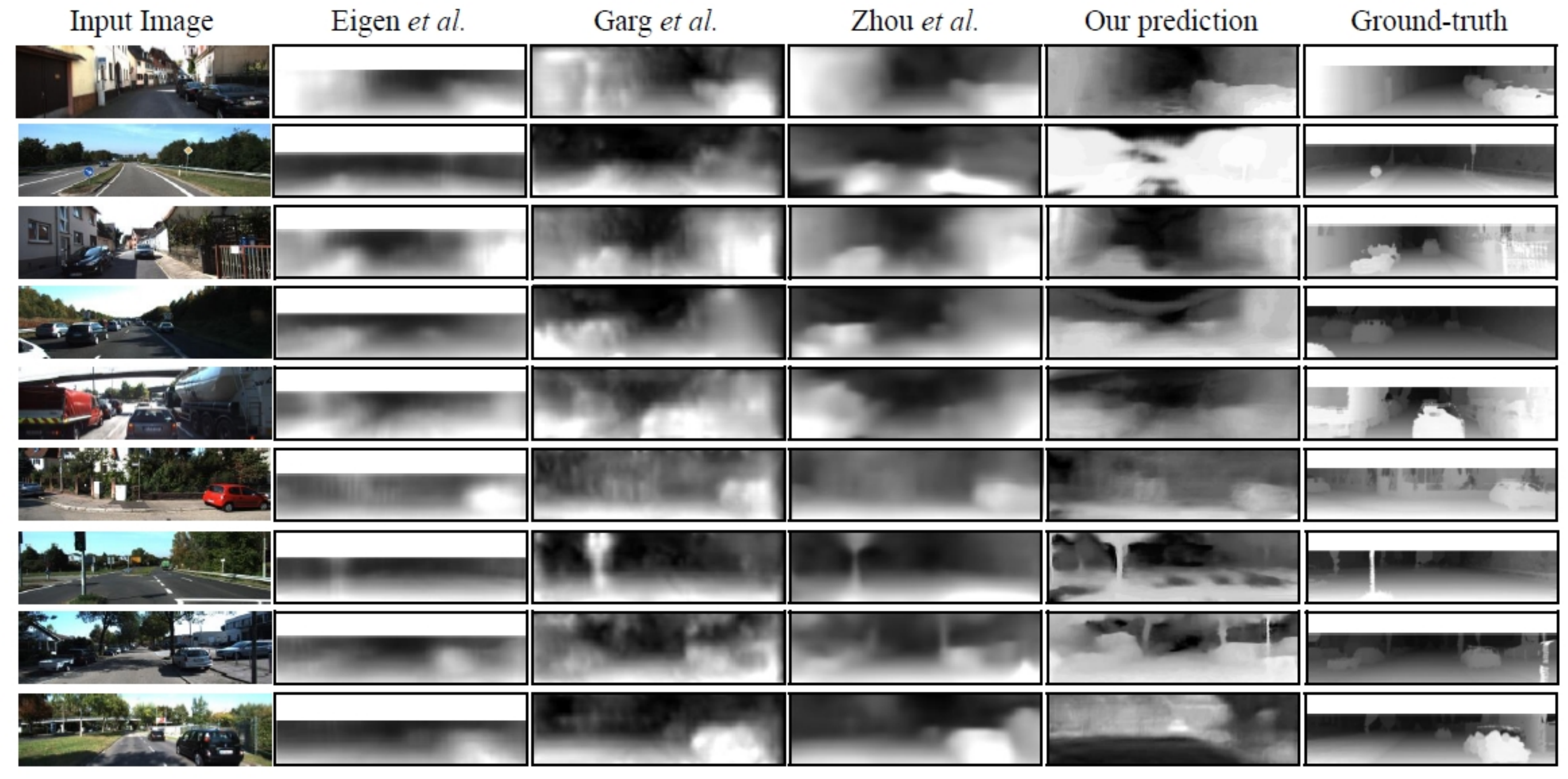}
	\caption{Comparison of single-view depth estimation between Eigen et al.\cite{eigen2014depth} (with ground-truth depth supervision), Garg et al. \cite{zhou2017unsupervised} (with
		ground-truth pose supervision), Zhou et al. \cite{zhou2017unsupervised} (unsupervised), and ours (unsupervised). The ground-truth depth map is interpolated from
		sparse measurements for visualization purpose.}
\end{figure} 

Compared with previous state-of-the-art method \cite{zhou2017unsupervised} using unsupervised mechanism, our method decreases the depth estimation error of nearly $13.5\%$. This validates that our learning framework
can effectively take advantage of the knowledge gained from the camera pose in traditional SLAM algorithms. We further compare the depth images obtained by our method with
the baseline methods. From Fig. 7, we can see that our results have no explicit difference with those of the supervised approaches. Furthermore, our method can even better represent the depth of the boundary information in some special scenes.
Fig. 8 visualizes the testing results of our method using the strategy of \cite{godard2017unsupervised} (first pre-train on the Cityscapes dataset and then fine-tune the model on the KITTI dataset).

\begin{figure}[htbp]
	\centering
	\includegraphics[width=0.9\textwidth]{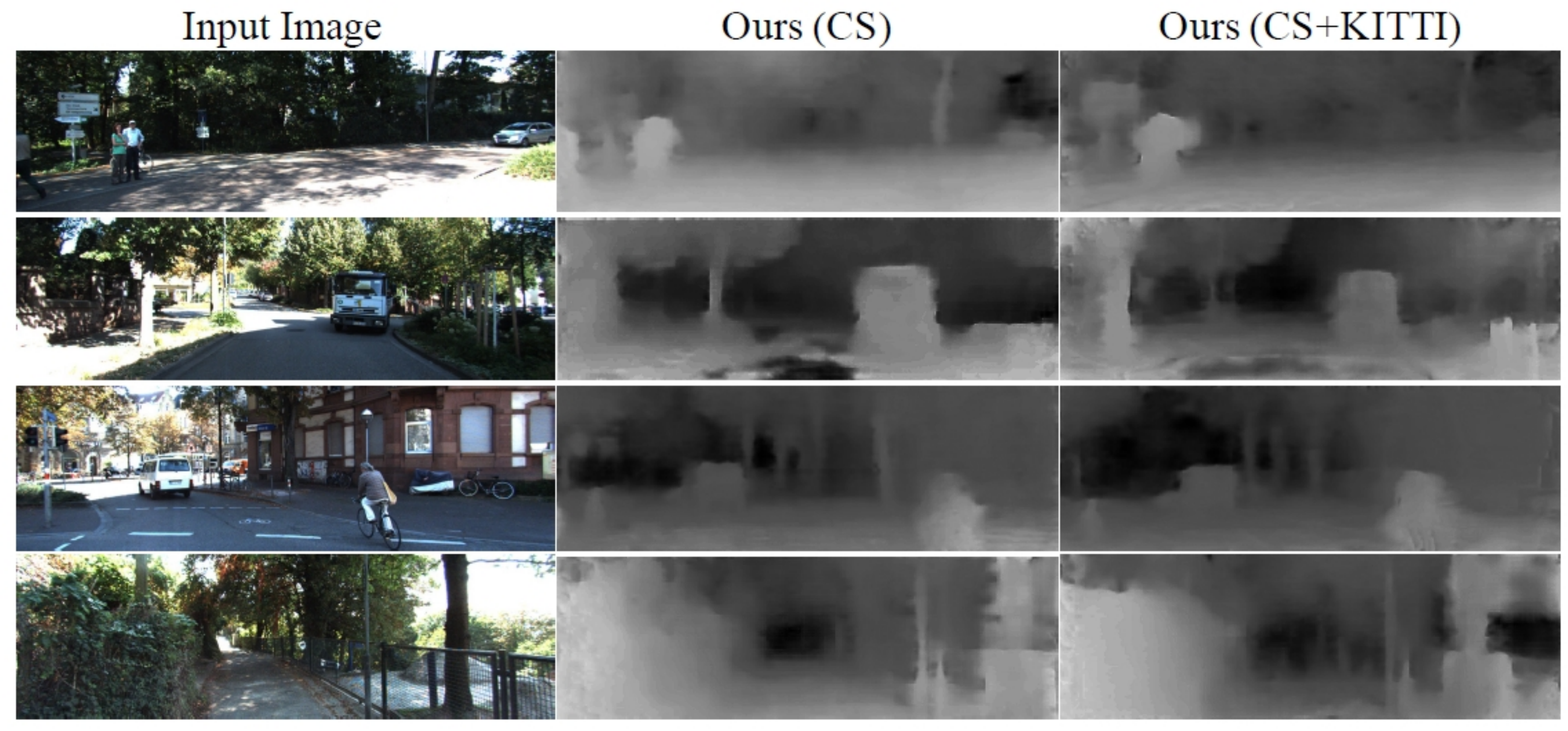}
	\caption{Comparison of single-view depth predictions on the KITTI
		dataset by our initial Cityscapes model and the final model (pretrained
		on Cityscapes and then fine-tuned on KITTI). The Cityscapes
		model sometimes ignores structural mistakes (e.g. roadside billboards
		and lamp posts) likely due to the domain gap between the two
		datasets.}
\end{figure}

In order to show the relationship between the length of the continuous frame
window and the performance of our system, we set the length of the window to 5 and repeat the experiments above. As shown in Table 2, the performance of 5-frame version version is better than that of the 3-frame version on all the metrics, which is consistent with our suppose. The benefit is attributed to the abundant transformation matrices between frames, which can further augment the supervised signals. 

\begin{table}[htbp] 
	\centering
	\caption{\label{tab:test}Results of different continuous frame window length versions of our system.}
\setlength{\tabcolsep}{2mm}{ 
\begin{tabular}{lccccc}
	\hline
	\multirow{2}{*}{Method}&
	\multirow{2}{*}{The length of the  }&
	\multicolumn{4}{c}{Error metric}\\
	\cmidrule(lr){3-6} 
	~&continuous frame window&Abs Rel&Sq Rel&RMSE&RMSE log\\
	\hline
	Zhou $et\ al.$ &3&0.208&1.768&6.856&0.283\\
	$\textbf{Ours}$ &3&$\textbf{0.180}$&$\textbf{1.510}$&$\textbf{6.349}$&$\textbf{0.256}$\\
	$\textbf{Ours}$ &5&$\textbf{0.176}$&$\textbf{1.455}$&$\textbf{5.940}$&$\textbf{0.248}$\\
	\hline
\end{tabular}}
\end{table}

\subsubsection{Make3D}

\begin{figure}[htbp]
	\centering
	\includegraphics[width=0.9\textwidth]{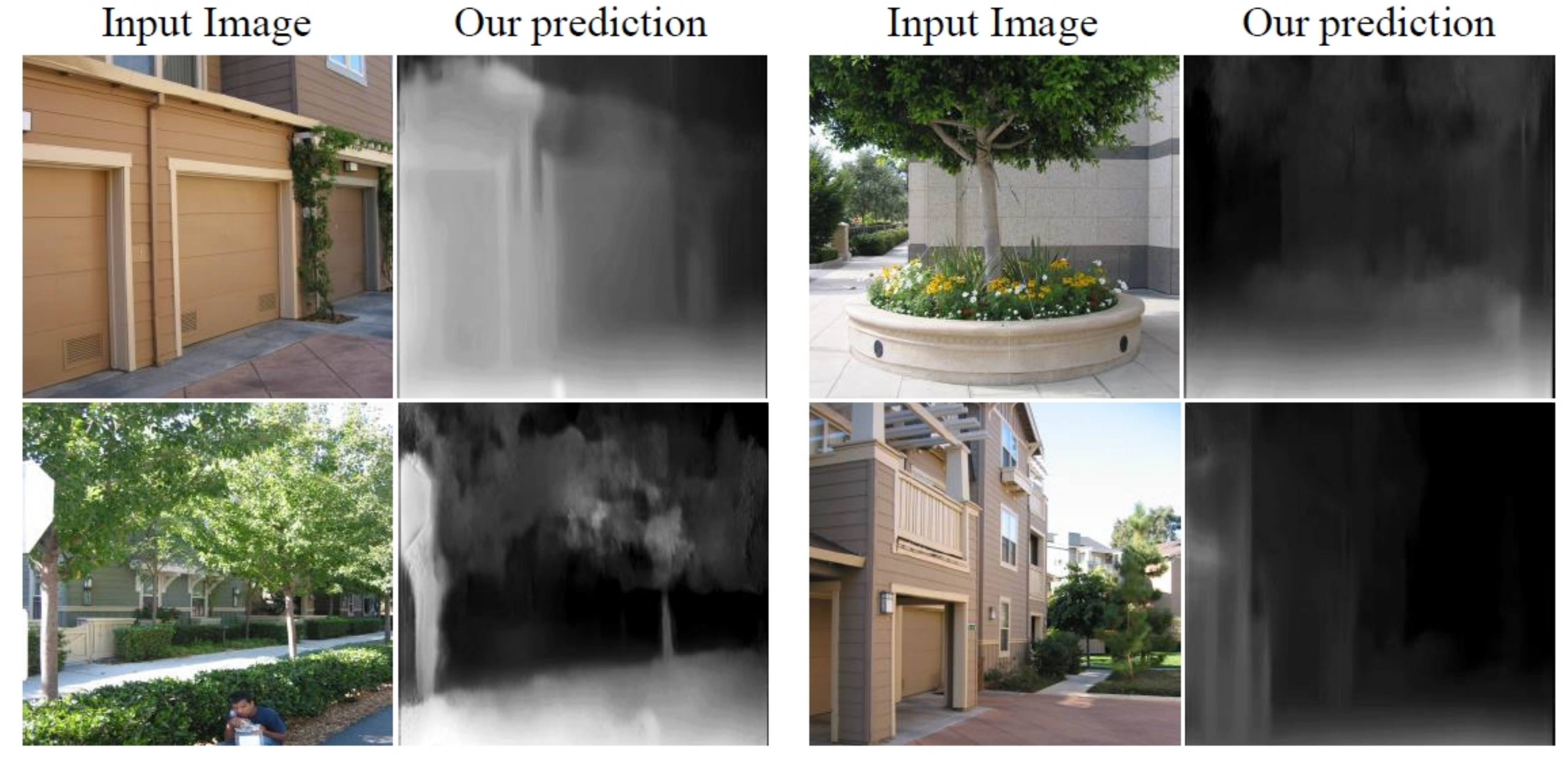}
	\caption{Our sample predictions on the Make3D dataset. Note that our
		model is trained on KITTI+Cityscapes only, and directly tested on
		Make3D.}
\end{figure}

In order to evaluate the generalization ability and the adaptability of our proposed method on the strong lighting and weak texture environments,
we choose Make3D \cite{saxena2009make3d} dataset for further experiments. In concrete, we pre-train our network on Cityscapes dataset and fine-tune on KITTI dataset. Then, we evaluate our model on
Make3D dataset, which contains abundant strong lighting and weak texture samples and maintains an explicit difference between the other two datasets. As shown in Table 3, the results on Make3D dataset are similar to those of KITTI dataset. In concrete, compared with the methods using depth supervision mechanism, our method achieves a better performance than \cite{eigen2014depth,liu2016learning }, but a little worse than \cite{godard2017unsupervised}, which indicates that our method maintains a satisfying generalization ability. Compared with the unsupervised method \cite{zhou2017unsupervised}, our method still 
achieves a better performance. We further visualize the sample
predictions of our method in Fig. 9 for a better illustration.

\begin{table}[htbp] 
	\centering
	\caption{\label{tab:test}Results on the Make3D dataset.} 
\setlength{\tabcolsep}{2mm}{
\begin{tabular}{lcccccc}
	\hline
	\multirow{2}{*}{Method}&
	\multicolumn{2}{c}{Supervision}&
	\multicolumn{4}{c}{Error metric}\\
	\cmidrule(lr){2-3} \cmidrule(lr){4-7} 
	~&Depth&Pose&Abs Rel&Sq Rel&RMSE&RMSE log\\
	\hline
	Train set mean&$\checkmark$&~&0.876&13.98&12.27&0.307\\
	Karsch $et\ al.$ &$\checkmark$&~&0.428&5.079&8.389&0.149\\
	Liu $et\ al.$ &$\checkmark$&~&0.475&6.562&10.05&0.165\\
	Laina $et\ al.$ &$\checkmark$&~&0.204&1.840&5.683&0.084\\
	Godard $et\ al.$ &$~$&\checkmark&0.544&10.94&11.76&0.193\\
	Zhou $et\ al.$ &$~$&~&0.383&5.321&10.47&0.478\\
	\hline
	$\textbf{Ours}$ &$~$&~&$\textbf{0.343}$&$\textbf{4.739}$&$\textbf{8.201}$&$\textbf{0.455}$\\
	\hline
\end{tabular}}
\end{table}

\subsection{Pose Estimation}

We choose ORB-SLAM \cite{mur2015orb} as the baseline method to illustrate the effectiveness of our proposed pose estimation network.
We follow the experimental settings in \cite{zhou2017unsupervised} and use the official KITTI odometry split method to guarantee a fair comparison.
The odometry benchmark is composed of 11 driving sequences with ground-truth
odometry. We choose the first 9 driving sequences (00-08) for
training and the last 2 driving sequences (09-10) for testing. The ground-truth odometry is used to evaluate our ego-motion
estimation performance and the length of the frame window is set to 5. We compare
our ego-motion estimation with two variants of monocular
ORB-SLAM algorithm. The first one is ORB-SLAM (full) which uses
all frames of the driving sequence to recover odometry. The second one is ORB-SLAM (short) which is lack of the loop closure and re-localization modules and maintains the same input setting as our system (5-frame snippets). Besides, the unsupervised method \cite{zhou2017unsupervised} is also selected as the baseline. 

\begin{figure}[htbp]
	\centering
	\includegraphics[width=0.9\textwidth]{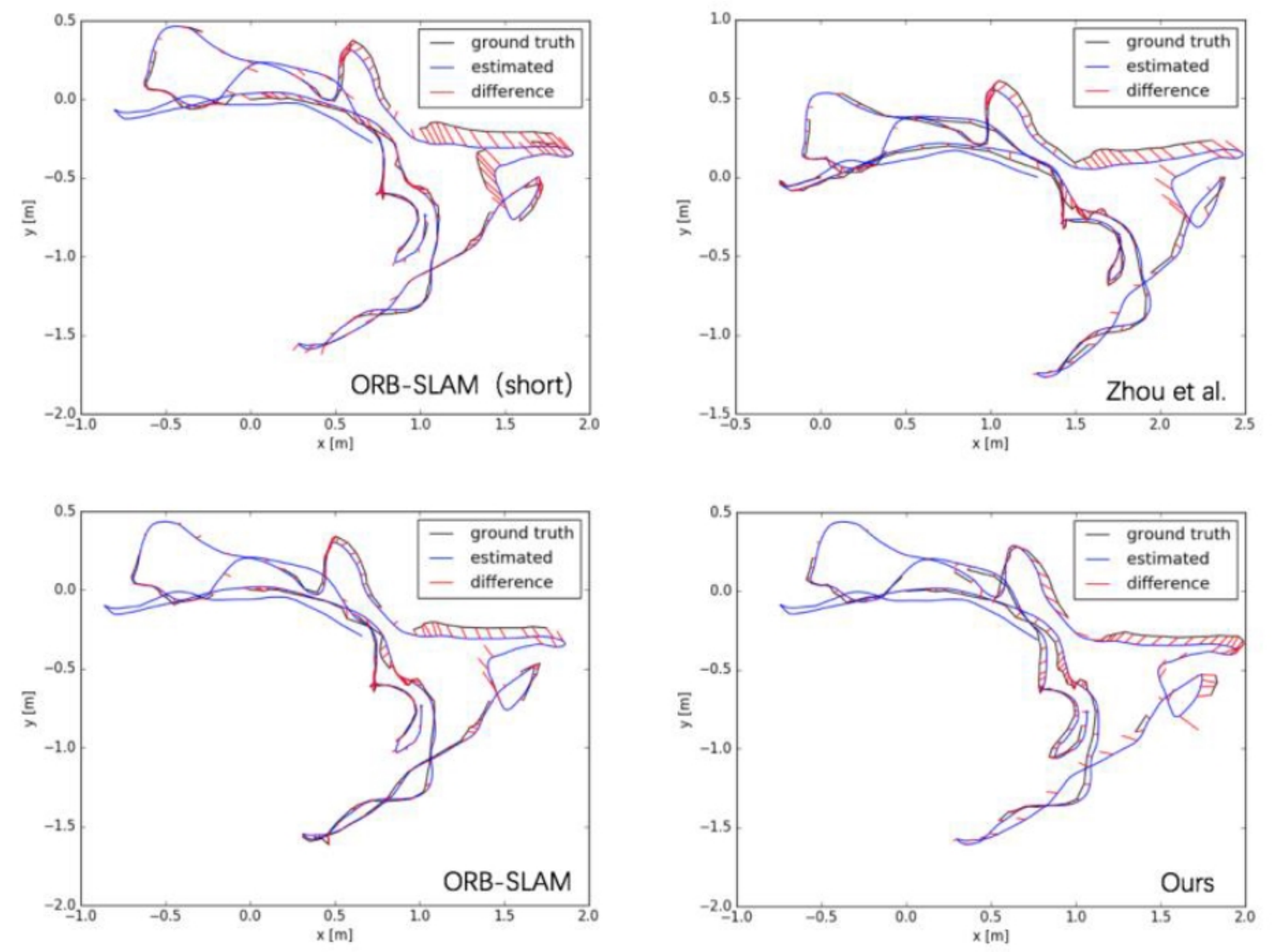}
	\caption{Pose estimation trajectories comparison.}
\end{figure}

Notably, due to the reason that
different methods have different scales, we optimize the scaling factor for the predictions made by each method to make
all the scaling factors consistent with the ground-truth. The Absolute Trajectory Error (ATE) of the ground-truth and the estimated trajectory is chosen for evaluation. All the methods are computed on 5-frame snippets except for ORB-SLAM (full). For
the ORB-SLAM (full) method, we break down the trajectory
of the full sequence into 5-frame snippets by adjusting the reference coordinate frame to the central frame of each snippet.

\begin{table}[htbp] 
	\centering
	\caption{\label{tab:test}Absolute Trajectory Error (ATE) on the KITTI odometry (lower is better).} 
\setlength{\tabcolsep}{7mm}{
\begin{tabular}{lcc}
	\hline
	Method & Seq.09 & Seq.10 \\
	\hline
	$\textbf{ORB-SLAM(full)}$ &$\textbf{0.014$\pm$0.008}$ &$\textbf{0.012$\pm$0.011}$\\
	\hline
	ORB-SLAM(short)&$0.064\pm0.141$&$0.064\pm0.130$\\
	Mean Odom&$0.032\pm0.026$&$0.028\pm0.023$\\
	ORB-SLAM(short)&$0.064\pm0.141$&$0.064\pm0.130$\\
	Zhou $et\ al.$&$0.021\pm0.017$&$0.021\pm0.017$\\
	$\textbf{Ours}$&$\textbf{0.017$\pm$0.008}$&$\textbf{0.015$\pm$0.017}$\\
	\hline
\end{tabular}}
\end{table}

As shown in Table 4, our approach performs comparably with
the ORB-SLAM (full) method, which utilizes the whole
image sequences for loop closure and re-localization to improve the pose estimation accuracy. The ATE value of our approach is about a quarter of
that acquired by ORB-SLAM (short). 
In future, it would be interesting to use our learned ego-motion
instead of the local estimation modules in monocular
SLAM systems. Meanwhile, our pose estimation outperforms the previous state-of-the-art unsupervised method \cite{zhou2017unsupervised}, which is conceptually similar to ours. Fig. 10 illustrates the pose estimation trajectories comparison between our method and the baseline methods.

\section{Evaluation in the Realistic Settings}

In this section, we first introduce the details on our platform for implementing the unsupervised learning-based depth estimation aided ORB-SLAM system using cloud robotic infrastructure. Then, we present the testing results of our system in various scenes. 

\subsection{Experimental Platform}

Our system is composed of two parts: the robot and the server from the perspective of hardware.  

The robot is deployed with multiple sensors, i.e., camera, ultrasonic radar and sound sensor. In our system, the robot interacts with the real-world settings by collecting the images from the RGB-D sensor and transmitting them to the server. The server mainly deals with the data saving and data processing tasks. In concrete, for our system, the depth estimation network and the pose estimation networks are both deployed in the server and the server will also processes the computing task and the simultaneous localization and mapping task. Fig. 11 illustrates the sparse point cloud image acquired by our system and the green lines stand for the pose trajectory predicted by our system.

\begin{figure}[htbp]
	\centering
	\includegraphics[width=0.9\textwidth]{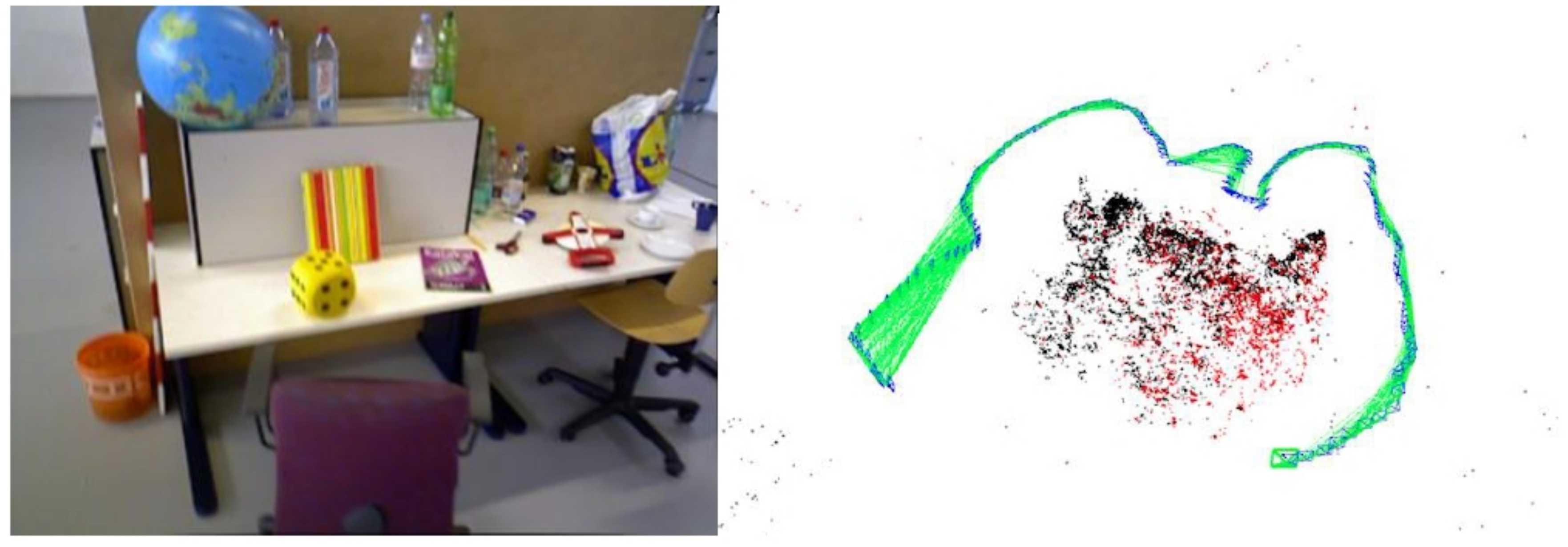}
	\caption{The sparse point cloud image acquired by our system on a indoor desk scene.}
\end{figure}

The depth estimation network and the pose estimation network deployed in the server are both well trained on KITTI, Cityscapes and Make3D datasets. We choose TUM RGBD dataset to evaluate the performance and the server uses NVIDA GeForce 1080P GPU for the processing tasks.	

\subsection{Experimental Results in Strong Lighting and Weak Texture Environments}

\begin{figure}[htbp]
	\centering
	\includegraphics[width=0.9\textwidth]{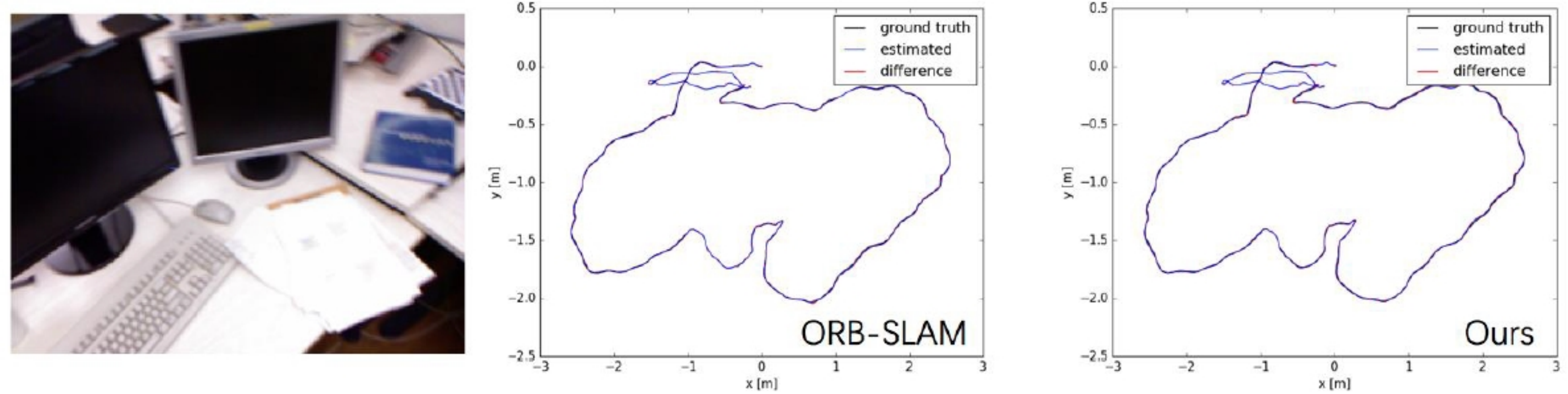}
	\caption{Illustration of pose estimation in a normal scene (desk).}
\end{figure}

\begin{figure}[htbp]
	\centering
	\includegraphics[width=0.9\textwidth]{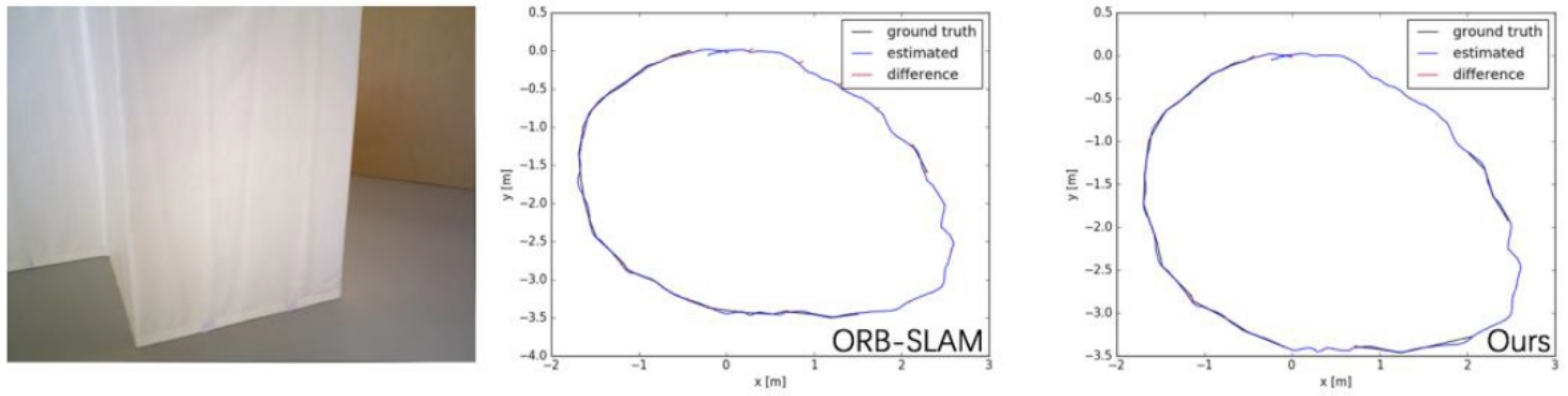}
	\caption{Illustration of pose estimation in a weak texture scene Fr3/nst.}
\end{figure}

\begin{figure}[htbp]
	\centering
	\includegraphics[width=0.9\textwidth]{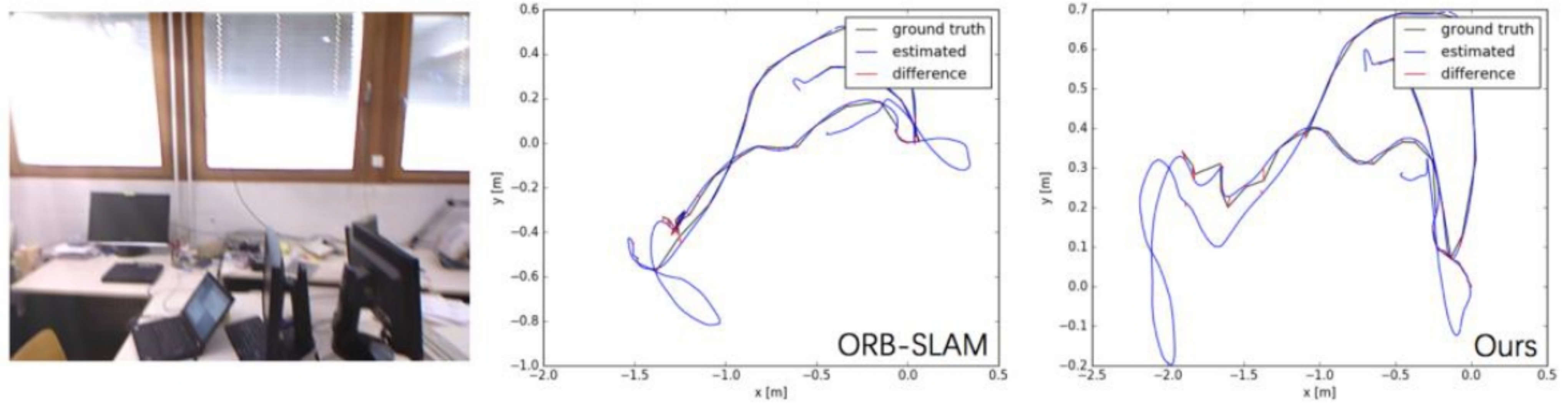}
	\caption{Illustration of pose estimation in a strong lighting scene Fr3/stf.}
\end{figure}

The standard for evaluating the performance of our system and the ORB-SLAM system is ATE, which is widely used for testing in the SLAM area. We compare the performances of the two methods in various scenes of TUM RGBD dataset. The results are reported in Table 5, the first 5 scenes are in the normal environments and the last 3 scenes are in the strong lighting or weak texture environments. We can see that in normal scenes, our system performs comparably to the traditional ORB-SLAM system. In the strong lighting and weak texture environments, our system achieves a better performance than that of the traditional ORB-SLAM system. For a better illustration, we report the pose estimation trajectories in normal, weak texture and strong lighting environments respectively (shown in Fig. 12, Fig. 13, and Fig. 14). 

\begin{table}[htbp] 
	\centering
	\caption{\label{tab:test}ATE Comparison of our method and traditional ORB-SLAM system in different scenes.} 
\setlength{\tabcolsep}{7mm}{
	\begin{tabular}{lcl} 
		\toprule 
		Scene & ORB-SLAM & Ours \\ 
		\midrule 
\emph{Fr1/desk} & 0.018490 & 0.017181 \\ 
\emph{Fr1/desk2} & 0.021034 & 0.021564 \\ 
\emph{Fr2/desk} & 0.011296 & 0.010978 \\
\emph{Fr1/room} & 0.061536 & 0.062283 \\
\emph{Fr2/office} & 0.010999 & 0.010901 \\
\emph{Fr2/stf} & 0.013178 & 0.012105 \\
\emph{Fr2/stn} & 0.012706 & 0.012294 \\
\emph{Fr2/nst} & 0.023507 & 0.022203 \\
		\bottomrule 
	\end{tabular}}
\end{table}

\subsection{Testing on the Speed for Initialization}

Due to the reason that our method optimizes the initialization process of the traditional ORB-SLAM system, we design a series of experiments to test the initialization speed of the two systems by recording the number of images used for initialization. 

\begin{figure}[htbp]
	\centering
	\includegraphics[width=0.9\textwidth]{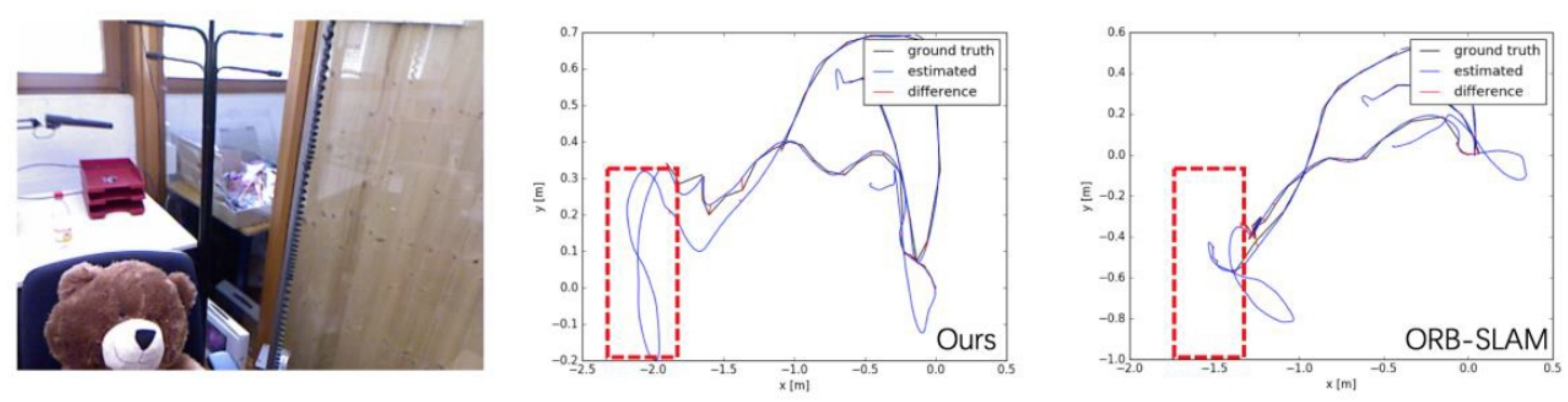}
	\caption{Initialization speed comparison in a strong lighting scene.}
\end{figure}

\begin{figure}[htbp]
	\centering
	\includegraphics[width=0.9\textwidth]{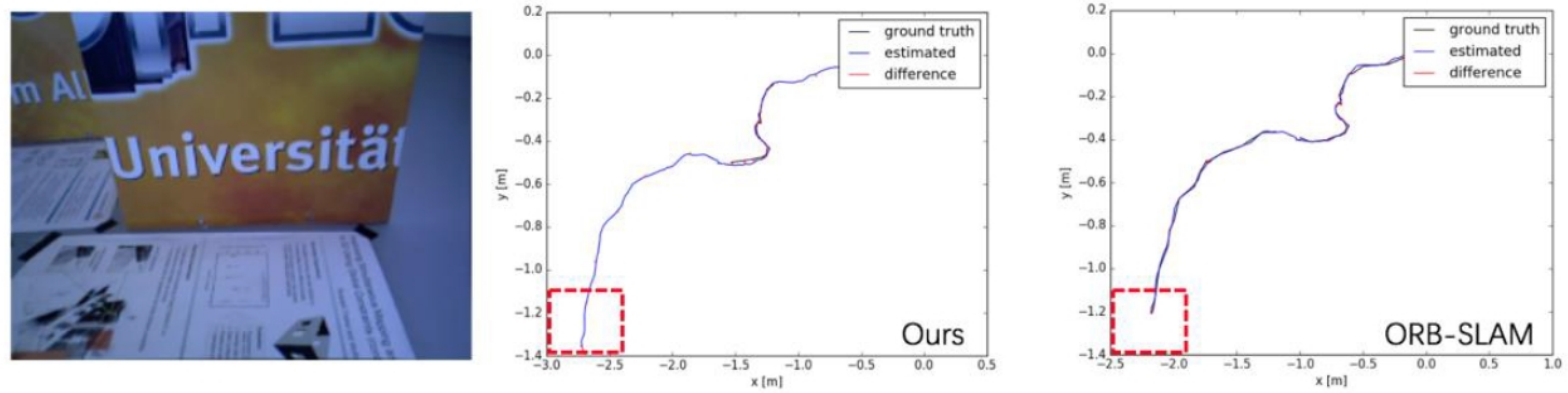}
	\caption{Initialization speed comparison in a weak texture scene.}
\end{figure}
Fig. 15 and Fig. 16 report the testing results in a strong lighting and weak texture  environments of the TUM RGBD dataset respectively. The lines in the red square stand for the estimated trajectories in the initialization process. We can see that the estimated trajectory of our method is longer than that of the traditional ORB-SLAM system in both the two situations, which indicates that our method can complete the initialization process and start to build the map in a faster speed. Table 6 reports the comparison on the number of images used in the initialization process. For the first 5 normal scenes, our system uses comparable number of images to the traditional ORB-SLAM system. However, towards the last 3 scenes in the strong lighting and weak texture environments, the number of images used by our system is explicitly less than that of the traditional ORB-SLAM system, which indicates the effectiveness of our approach. 

\begin{table}[htbp] 
	\centering
	\caption{\label{tab:test}Comparison on the number of images used in the initialization process.} 
	\setlength{\tabcolsep}{7mm}{
		\begin{tabular}{lcl} 
			\toprule 
			Scene & ORB-SLAM & Ours \\ 
			\midrule 
			\emph{Fr1/desk} & 4 & 2 \\ 
			\emph{Fr1/desk2} & 5 & 2 \\ 
			\emph{Fr2/desk} & 4 & 4 \\
			\emph{Fr1/room} & 7 & 2 \\
			\emph{Fr2/office} & 6 & 5 \\
			\emph{Fr2/stf} & 38 & 10 \\
			\emph{Fr2/stn} & 145 & 84 \\
			\bottomrule 
	\end{tabular}}
\end{table}

\section{Conclusion}

We present an unsupervised learning framework for single-view depth and ego-motion estimation. The proposed method exploits the pose estimation method to enhance the supervised signal and add training constraints for the task of monocular depth and camera motion estimation. The system is trained on unlabeled videos and performs
comparably to approaches that require ground-truth depth or
pose for training. Furthermore, our method outperforms the previous state-of-the-art unsupervised learning method by $13.5\%$ on KITTI dataset. Finally, we successfully exploit our unsupervised learning framework to assist the traditional ORB-SLAM system when the initialization module of ORB-SLAM method could not match enough features. Experiments have shown that our method can significantly accelerate the initialization process of traditional ORB-SLAM system and effectively improve the accuracy on environmental mapping in strong lighting and weak texture scenes.

\subsubsection*{Acknowledgments.} This work was supported by the National Natural Science Foundation of China (grant
numbers 61751208, 61502510, and 61773390), the Outstanding Natural Science Foundation of Hunan Province
(grant number 2017JJ1001).

\bibliographystyle{splncs03}
\bibliography{mybib}

\end{document}